# KARMA: Augmenting Embodied AI Agents with Long-and-short Term Memory Systems

**Zixuan Wang[1,6], Bo Yu[2,*], Junzhe Zhao[3], Wenhao Sun[4,6], Sai Hou[5], Shuai Liang[4],**
Xing Hu[4], Yinhe Han[4], Yiming Gan[4,*]

## Abstract

Embodied AI agents responsible for executing interconnected, long-sequence household tasks often face difficulties with in-context memory, leading to inefficiencies and errors in task execution. To address this issue, we introduce KARMA, an innovative memory system that integrates long-term and short-term memory modules, enhancing large language models (LLMs) for planning in embodied agents through memory-augmented prompting. KARMA distinguishes between long-term and short-term memory, with long-term memory capturing comprehensive 3D scene graphs as representations of the environment, while short-term memory dynamically records changes in objects' positions and states. This dual-memory structure allows agents to retrieve relevant past scene experiences, thereby improving the accuracy and efficiency of task planning. Short-term memory employs strategies for effective and adaptive memory replacement, ensuring the retention of critical information while discarding less pertinent data. Compared to state-of-the-art embodied agents enhanced with memory, our memory-augmented embodied AI agent improves success rates by $1.3\times$ and $2.3\times$ in Composite Tasks and Complex Tasks within the AI2-THOR simulator, respectively, and enhances task execution efficiency by $3.4\times$ and $62.7\times$. Furthermore, we demonstrate that KARMA's plug-and-play capability allows for seamless deployment on real-world robotic systems, such as mobile manipulation platforms.Through this plug-and-play memory system, KARMA significantly enhances the ability of embodied agents to generate coherent and contextually appropriate plans, making the execution of complex household tasks more efficient. The experimental videos from the work can be found at https://youtu.be/4BT7fnw9ehs. Our code is available at https://github.com/WZX0Swarm0Robotics/KARMA/tree/master.

[1]Institute of Automation, Chinese Academy of Science (CASIA)
[2]Shenzhen Institute of Artificial Intelligence and Robotics for Society
[3]Alibaba Group, Hangzhou, China
[4]Institute of Computing Technology, Chinese Academy of Sciences (ICT, CAS)
[5]Beijing Institute of Technology
[6]University of Chinese Academy of Sciences
* Corresponding author.

## 1 Introduction

Robotic applications are evolving towards longer and more complex tasks. Using an LLM as its core planning module can effectively decompose long and complex tasks into multiple short and fixed movements (Choi et al., 2024; Sarch et al., 2024; Chen et al., 2023b; Vemprala et al., 2023; Rana et al., 2023; Brohan et al., 2022, 2023; Belkhale et al., 2024), increasing the success rate.

Yet, simply equipping an embodied agent or a robot with an LLM is not enough. Take indoor household tasks as an example, they usually require a sequence of interrelated instructions where later ones have strong or weak dependencies on previous ones. When the amount of in-context examples and task descriptions necessary to cover the task constraints increases, even advanced models like GPT-4o can blur critical details, such as the location of a previously used object. Thus, there is a growing need to enhance the power of LLMs with "memory-augmented prompting" (Sarch et al., 2023a; Lewis et al., 2020; Mao et al., 2020).

We introduce KARMA, a plug and play memory system tailored for indoor embodied agents. The memory system comprises both long-term memory, represented as a non-volatile 3D scene graph, and volatile short-term memory, which retains immediate information about objects encountered during instruction execution. The memory system allows agents to accurately recall the positions and states of objects during complex household tasks, reducing task redundancy and enhancing execution efficiency and success rates.

On top of the memory system design, we propose to effectively maintain the contents of the memory given the capacity constraints. Specifically, we use the metric hit rate that measures how often a memory recall requirement is satisfied. We demonstrate that a higher hit rate indicates an improved replacement policy and enhanced system performance. Using this metric, we propose replacing the least recently used (LRU) unit whenever a new unit needs to be incorporated into a full memory. Our findings show that this approach achieves a higher hit rate compared to a naive first-in-first-out policy.

In summary, the paper makes following contributions to the community:

- We tailor a memory system for household embodied agents, which combines a long-term memory module and a short-term memory module. We also propose the way of recalling from both modules and feeding it to the LLM planner.

- We propose to use hit rate as the metric of evaluating the effectiveness of memory replacement mechanism and present to always replace the least frequently used unit with the new unit.

- We evaluate the memory-augmented LLM planner in the simulated household environments of ALFRED (Shridhar et al., 2021) and AI2-THOR (Kolve et al., 2017). Additionally, we conduct experiments by deploying KARMA on real robotic systems. The results shows significant improvements in the efficiency and accuracy of embodied agents performing long-sequence tasks.

## 2 Related Work

### 2.1 LLM for Robotics

Large language models have been widely used in robotic applications (Huang et al., 2022; Ahn et al., 2022) due to their impressive generalization abilities and common-sense reasoning capabilities (Brown et al., 2020; Madaan et al., 2022; Achiam et al., 2023). In most cases, LLMs replace the task planning and decision making modules in traditional robotic computing pipeline. Most robotic applications now encode sensor inputs into the format of LLM-accepted tokens and use LLMs to generate the next instructions, which further connect to robots through predefined skills or basic movements across different degrees of freedom (Ahn et al., 2022; Jin et al., 2023; Wu et al., 2023a,c).

### 2.2 Memory-Augmented Prompting of LLM-Based Agent

Using LLMs as task planner for robots face the challenge of accurately retaining information across multiple interdependent tasks. Thus, augmenting LLM-based agents with different forms of memory is a common approach in role-playing games (Shao et al., 2023; Li et al., 2023a; Wang et al., 2023e; Zhou et al., 2023; Zhao et al., 2023), social simulations (Kaiya et al., 2023; Park et al., 2023; Gao et al., 2023; Li et al., 2023b; Hua et al., 2023), personal assistants (Zhong et al., 2024; Modarressi et al., 2023; Lu et al., 2023; Packer et al., 2023; Lee et al., 2023; Wu et al., 2023b; Hu et al., 2023; Liu et al., 2023; Liang et al., 2023), open-world games (Wang et al., 2023a; Zhu et al., 2023; Wang et al., 2023f; Yan et al., 2023), code generation (Tsai et al., 2023; Chen et al., 2023a; Qian et al., 2023; Li et al., 2023b; Zhang et al., 2024b), recommendations (Wang et al., 2023d,c; Zhang et al., 2024a), and domain-specific expert systems (Wang et al., 2023b; Yang et al., 2023; Zhao et al., 2024b).

The definition and formats of the memory is distinctive in different works. Historical actions (Park et al., 2023), thoughts (Liu et al., 2023), contexts (Liang et al., 2023; Packer et al., 2023) are explored. Different memory management mechanisms are also designed and evaluated. For example, agents can simply use text indexing to match relevant memory; the memory recall and management can also be much more complicated, involving text embedding, semantic retrieval(Zhao et al., 2024a) and Graph RAG(Edge et al., 2024).

In the field of embodied agents, much of the research (Kagaya et al., 2024; Zhang et al., 2023; Sarch et al., 2023a; Wang et al., 2023f) focuses on storing and recalling past experiences, allowing agents to learn from previous interactions and make more informed decisions. Research (Kim et al., 2023) uses short-term memory to maintain and continuously track the positions of objects through semantic labels. Additionally, other studies utilizes structured maps as long-term memory, enabling agents to more efficiently locate places or objects in vision-language navigation tasks(Zhan et al., 2024; Chiang et al., 2024). However, these approaches fail to address challenges such as hallucinations or memory inconsistencies that often arise long-sequence task planning with LLMs. Furthermore, integrating memory mechanisms into LLMs remains at a preliminary stage, particularly regarding memory saving and updating mechanisms. For example, saving everything permanently can result in unaffordable storage requirements, while refreshing the memory every time agents restart will lose any long-term capability. Additionally, the decision of which memory unit to replace remains unsolved. Most approaches use either a forgetting curve (Zhong et al., 2024) or the simple first-in-first-out principle (Packer et al., 2023) without detailed discussions on context-specific updates.

Our work addresses these limitations by incorporating a tailored memory framework for embodied AI agents. This system includes long-term memory in the form of a 3D scene graph representing static objects and short-term memory for instant information about recent activities. This long-short memory approach helps the agent better understand its environment and recent actions. Various exit and update mechanisms are discussed to maintain effectiveness even under fixed memory capacity, providing a comprehensive solution for long sequential tasks in household environments.

# 3 Method

We describe the methodology in this section, with start on elaborating the problem setup (Sec. 3.1), Sec. 3.2 gives an overview of the framework and Sec. 3.3 and Sec. 3.4 reveals the long-term and short-term memory design. We wrap Sec. 3.6 with the novel memory exit and replacement mechanism.

## 3.1 Problem setup

Although generalizable, our work focuses on indoor environment where users send instructions to an agent to perform a series of tasks, $H = I_{t_0}, I_{t_1}, \ldots, I_{t_N}$. These tasks are typically related in terms of both time and order of completion. For instance, if the agent is asked to prepare a salad, it must first wash an apple ($I_{t_0}$) and cut it ($I_{t_1}$), then repeat the process with a tomato ($I_{t_2}$,$I_{t_3}$), and finally place the ingredients into a bowl and mix them. During this process, an large volume of high-dimensional data is incorporated through various sensors, such as the agent's location and the position and status of different objects. Even when equipped with a large language model as its planner, the agent may lose track of its tasks and need to re-explore the environment, which motivates our work to customize a memory system to augment the agent.

In this paper, we use $S \in \{S_{\text{manipulation}} \cup S_{\text{navigation}}\}$ to represent the set of skills that the agent can perform, which should be executed by a LLM through pre-defined APIs. The instruction $I$ can be further decomposed into an ordered set of $K$ sub-tasks, $T = \{T_1, T_2, \ldots, T_K\}$, where $K$ represents the sequence of sub-tasks over time.

## 3.2 Overview

KARMA is a memory system tailored for embodied AI agents, incorporating memory design, recall using context embedding with a pre-trained LLM and an accurate replacement policy. Specifically, we design two memory modules: long-term memory and short-term memory. The long-term memory comprises a 3D scene graph (3DSG) representing static objects in the environment, while the short-term memory stores instant information about used or witnessed objects. The long-term memory aids the agent in better understanding the environment, and the short-term memory helps the agent understand its recent activities. Due to fixed memory capacity, we also discuss various exit and update mechanisms. Fig. 1 provides an overview of our work.

## 3.3 Long-Term Memory Design

Long-term memory is large in size, non-volatile, and task-irrelevant. It should be built incrementally and updated infrequently. This type of memory is designed to store static information that remains constant over extended periods, such as the layout of the environment and the positions of immovable objects. In the context of an indoor agent, semantic maps serve as an appropriate carrier for it.

In many forms of semantic maps, KARMA uses a 3D scene graph to represent the environment. The main reason we choose a 3DSG instead of 2D semantic maps or voxel grids is that 3DSG offers a more accurate and comprehensive representation of the environment and features a topological structure, which is essential for tasks that require precise navigation and manipulation. Also, even a state-of-the-art multi-modality LLM has difficulties understanding the geographic relationships from a 2D semantic map, while a 3DSG display it explicitly.

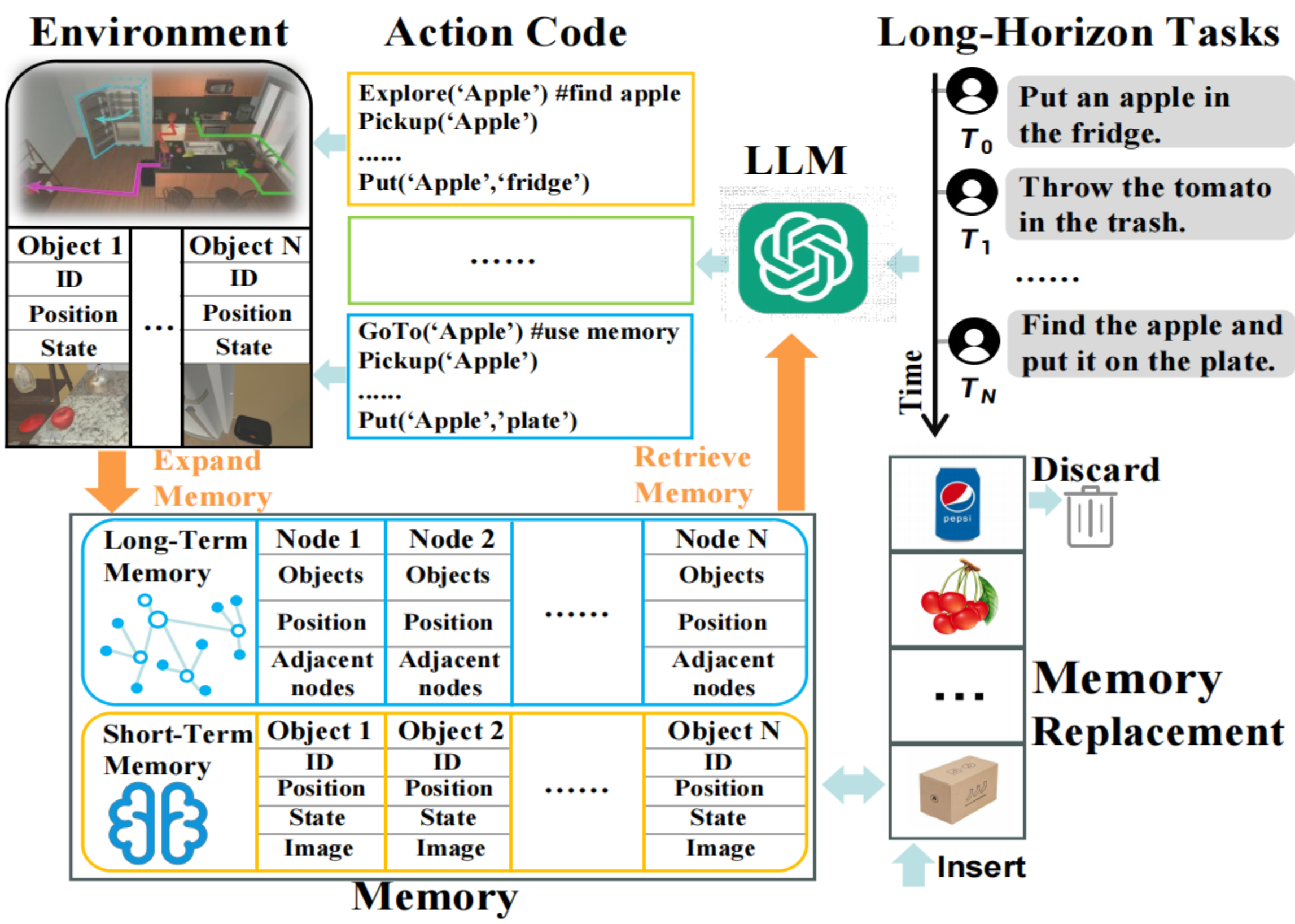

Fig. 1: KARMA's architecture, with a LLM as the core planner, a long-term and a short-term memory, and corresponding recall and replacement mechanisms.

The 3DSG utilizes a hierarchical structure encompassing floors, areas, and objects, not only capturing the spatial relationships and attributes of objects but also leveraging the benefits of a topological graph. This structure is particularly advantageous when expanding the map to represent the environment, as its sparse topological nature effectively mitigates the impact of accumulated drifts compared to dense semantic maps. Thus, 3DSG is better suited to meet the navigation needs in unknown environments. The construction process of the 3DSG is similar to existing works (Rosinol et al., 2021; Armeni et al., 2019; Rana et al., 2023), as illustrated in Figure 2. We establish and manage a hierarchical topological graph $G = (V, E)$, where the set of vertices $V$ is composed of $V_1 \cup \ldots \cup V_k$, with $k = 3$, Each $V_k$ represents the set of vertices at a particular level of the hierarchy. The area nodes, $V_2 = \{V_2^1, V_2^2, \ldots, V_2^N\}$, are evenly distributed across the reachable regions in the indoor environment, with their world coordinates acquired through a simulator. If two area nodes are navigable to each other, an edge is established between them. For each area node, we detect the types and additional information of objects within a certain radius, using data acquired through a simulator. In real-world applications,this object detection can be performed using methods such as Faster R-CNN. The detected immovable entities are then assigned as object nodes to their respective area nodes. These object nodes encode detailed attributes such as volume and 3D position.

In our framework, the agent gradually builds and maintains a 3DSG as it explores the indoor environment. The graph remains unchanged unless the indoor environment change. When being used by the planner, we transform the 3DSG into a topological graph and serialized it into a text data format that can be directly parsed by a pre-trained LLM. An example of a single area node from the 3DSG is as follows: {name: node_1, type: Area, contains: [bed, table, window, ...], adjacent nodes: [node_2, node_8], position: [2.34, 0.00, 2.23]} with edges between nodes captured as $\{\text{node_1} \leftrightarrow \text{node_2}, \text{node_1} \leftrightarrow \text{node_8}\}$.

Our design and use of long-term memory aim to provide accurate geometric relationships within the indoor environment. With this information, the agent is able to reduce the cost for repetitive environment exploration by allowing the addition or deletion of nodes through topological relationships, thus updating the environment representation seamlessly. This approach effectively avoids the drift errors typically caused by loop closure detection in traditional SLAM methods, and it minimizes the need for extensive place recognition processes, saving computational resources, storage, and time.

Moreover, long-term memory enhances the agent's ability to make informed decisions based on a comprehensive understanding of the environment. This capability is particularly useful for planning complex, multi-step tasks. By accessing detailed and persistent environmental data, the agent can predict potential obstacles and plan its actions more effectively, thereby improving both task completion success rates and execution efficiency. Also, the 3DSG is updated when the indoor environment changes, capturing the up-to-date information.

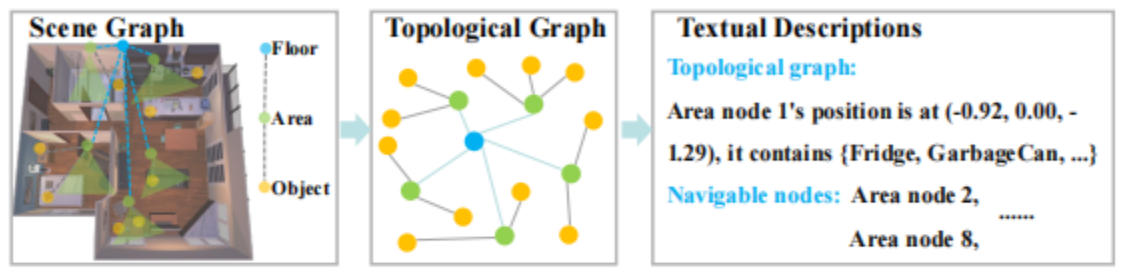

Fig. 2: Transforming 3D scene graphs into prompts.

### 3.4 Short-Term Memory

Short-term memory is small, volatile, and frequently updated. It is refreshed every time the agent starts and provides instant memorization of recently used objects and their status during task execution. This ensures that the same objects or relevant information are readily available for subsequent tasks.

Among all the information the agent captures during tasks, vision data is relied upon, as it provides the highest information density compared to other sensor inputs. After capturing an image, we use a vision language model (VLM) to analyze the image and extract the state of the object of interest (OOI). This process is task-specific, meaning the VLM is fed both the task and the image to handle multiple objects in the image. Subsequently, the world coordinates (acquired through a simulator), the state (generated by the VLM), and the raw image form a memory unit in the short-term memory, akin to a line of data in a cache. Finally, a multi-modality embedding model converts the memory unit into a vector for later recall.

We use an example to illustrate the design of KARMA's short-term memory. Given a task asking the agent to 'wash an apple and place it in a bowl,' the agent will memorize the coordinates of the apple and its state (cleaned) at the end. If a subsequent task asks the agent to 'bring an apple,' KARMA will retrieve the apple's memory from short-term memory, include it in the prompt, and query the LLM to generate a more efficient task plan. This saves the agent from exploring the kitchen to find the apple, reduces interactions with the LLM, and speeds up the process.

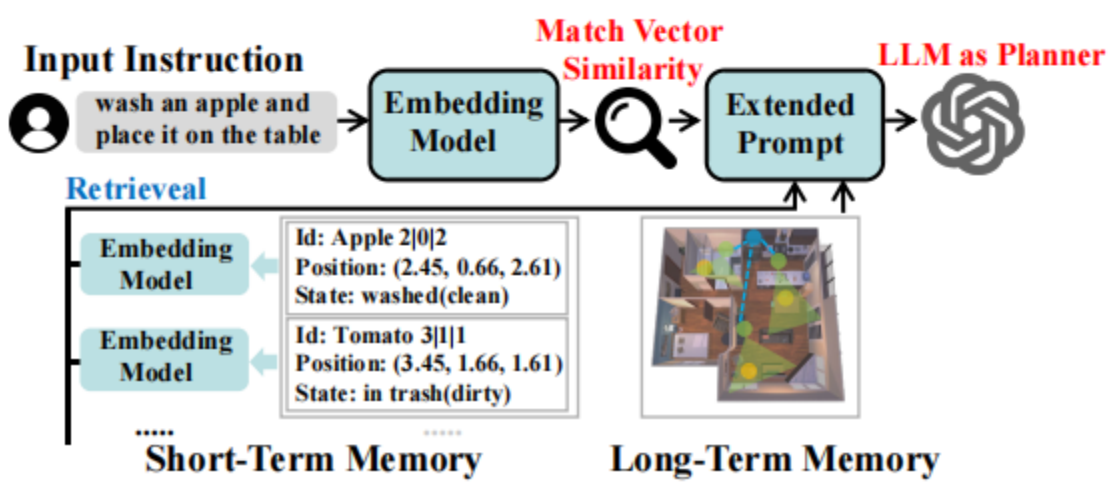

Fig. 3: Recalling long-term and short-term memory

### 3.5 Planner

KARMA utilizes two memory modules to augment the planning process, in order to achieve higher success rates and lower costs. We first decompose a given instruction $I$ into a sequence of subtasks or skills $S \in \{S_{\text{manipulation}} \cup S_{\text{navigation}}\}$. These skills include basic agent actions such as `Explore()` and `Openobject()`, which are pre-programmed. The planner call the skills through a set of APIs (Kannan et al., 2023; Sarch et al., 2023a). More details of APIs are provided in Apdx. D.

KARMA's planner uses both long-term and short-term memory when interacting with the LLM. As mentioned earlier, the entire long-term memory is directly serialized into the prompt, while only one unit of the short-term memory can be selected. KARMA uses vector similarity to select from the entire short-term memory. Each short-term memory is embedded into a set of vectors using a pre-trained embedding model. For the current instruction $I$, KARMA retrieves the top-K most similar memories—those with the smallest cosine distance to the embedding of the input instruction $I$. The corresponding text content of these memories is then added as context to the LLM prompt.

We show an example prompt in Apdx. A. It includes the action code for the basic skills S (parameterized as Python functions), examples of task decomposition, the input instruction $I$, and the retrieves short-term memory and long-term memory. The LLM is tasked with generating action code based on the parameterized basic skills S.

### 3.6 Memory Replacement

Unlike long-term memory that can be stored in non-volatile storage, short-term memory has a fixed capacity and can easily become full. An effective short-term replacement policy ensures it remains highly relevant to subsequent tasks.

**Hit rate.** We use memory hit rate to evaluate the effectiveness of memory replacement policies. This metric is defined as the ratio of the number of times the required memory units are found in short-term memory to the total number of queries. It is widely used in evaluating cache replacement policies(Einziger and Friedman, 2014), with higher values indicating better performance.

**First-In-First-Out (FIFO).** The FIFO replacement policy is the most straightforward. It manages memory units as a queue. When the queue is full and a new memory unit needs to be added, the earliest entry will be removed from the queue.

We improve the FIFO policy to better suit our application by adding a merging option. When a new memory unit needs to join the queue and the queue is full, we first check the object's ID in all memory units in the queue. If the same ID exists, the new unit will replace the old one with the same object's ID, instead of replacing the oldest unit.

**Least Frequently Used.** A more complex yet accurate replacement policy is Least Frequently Used (LFU). The design principle of LFU is based on the usage frequency of each memory unit. Whenever a new memory unit needs to join, the existing unit with the lowest usage frequency is replaced. This results in a high hit rate, as the memory retains frequently-used units. Since perfect LFU is not feasible, we use an approximate method called W-TinyLFU.

W-TinyLFU maintains two segments of memory: a main segment and a window segment. The main segment is organized in a two-segment Least Recently Used (LRU) manner, containing a protection segment and an elimination segment. Units in the protection segment are the safest; even if they are picked for replacement, they first move to the elimination segment.

Every time a unit needs to join the memory, it enters the window segment first. When the memory is full and a unit needs to be evicted, a comparison occurs among all units in the window segment and the elimination segment. The memory then selects the unit whose eviction would minimally impact the overall usage frequency and evicts it.

W-TinyLFU uses counting Bloom filters (Luo et al., 2018) as the basic data structure to count the usage of memory units. To keep frequency statistics fresh, W-TinyLFU applies a reset method. Each time a memory unit is added, a global counter is incremented. When the counter reaches a threshold $W$, all counters are halved:$c_i \leftarrow \frac{c_i}{2}$.

## 4 Experiments

We discuss the setup Sec. 4.1 and metrics Sec. 4.2 first, followed by extensive experiments. This includes success rate and efficiency (Sec. 4.3), different replacement policies (Sec. 4.4), ablation study (Sec. 4.5) and real-world deployment(Sec. 4.6).

### 4.1 Experimental Setup and Metrics

**Experimental Settings.** We use the widely-adopted AI2-THOR simulator(Kolve et al., 2017) for evaluation. The simulator's built-in object detection algorithm provided the label of objects and their relevant information for both long-term and short-term memory. Additionally, we employ OpenAI's text-embedding-3-large model as the embedding model for memory recall.

**Baseline.** To our best knowledge, most current methods using LLMs for task planning are very similar with LoTa-Bench(Choi et al., 2024). It provides a prompt that includes a prefix, in-context examples to the LLM, and then the LLM calculates the probabilities of all executable skills based on this prompt and selects the skill from skill sets most likely to complete the task. We also use it as our baseline. Additionally, we optimize the efficiency and success rate of planning and executing tasks in LoTa-Bench by referring to the skill sets configurations and selection described in SMART-LLM(Kannan et al., 2023).

**Dataset.** The dataset construction utilizes tasks from the ALFRED benchmark(Shridhar et al., 2021). By extracting its typical tasks and reorganizing them into long sequence tasks that align with everyday human needs, we ensured a more accurate assessment. More details of the dataset are provided in supplementary material.

This new dataset, ALFRED-L, includes 48 high-level instructions that detail the length, relevance, and complexity of sequential tasks. Additionally, it provides corresponding AI2-THOR floor plans to offer spatial context for task execution. We also include the ground truth states and corresponding location of objects after the completion of each subtask. This ground truth is used as symbolic goal conditions to determine whether the tasks are successfully completed. For example, conditions such as heated, cooked, sliced, or cleaned are specified. Our dataset comprises three task categories:

Simple Tasks have multiple unrelated tasks. The agent is assumed to perform sequential tasks with a length of less than five, without requiring specific memory to assist in task completion.

Composite Tasks include highly related tasks. These tasks involve multiple objects, and the agent needs to utilize memories generated from previous related tasks to execute subsequent subtasks.

Complex Tasks consist of multiple loosely related tasks. Some of these tasks involve specific objects, while others involve vague object concepts. For example, the agent be instructed to wash an apple($I_{t_0}$) and cut it($I_{t_1}$), then to place a red food on the plate($I_{t_2}$).

ALFRED-L comprises 15 tasks categorized as simple tasks, 15 tasks as composite tasks, 18 tasks as complex tasks.

Additionally, we use another dataset to better assess the performance of the memory replacement mechanism. The new dataset, ALFWorld-R, consists of long-sequence tasks $H = \{I_{t_0}, I_{t_1}, ..., I_{t_N}\}$, with each task $I_{t_i}, i \in \{0, 1, 2, ..., N\}$ in the sequence randomly selected from tasks in ALFRED.

### 4.2 Evaluation Metrics.

**Success Rate (SR)** is the percentage of tasks fully completed by the agent. A task is considered complete only when all subtasks are achieved.

**Memory Retrieval Accuracy (MRA)** is a binary variable determines if related memory can be successfully retrieved.

**Memory Hit Rate (MHR).** The definition is the same as the hit rate described in Sec. 3.6.

**Reduced Exploration (RE).** This metric measures the effectiveness of the system in reducing unnecessary exploration attempts. $RE = \frac{E_{\text{reduced}}}{E_{\text{total}}}$, where $E_{\text{total}}$ is the total number of exploration attempts, $E_{\text{reduced}}$ is the number of exploration attempts that were reduced.

**Reduced Time (RT).** This metric measures the proportion of time saved by reducing unnecessary actions during task execution. $RT = \frac{T_{\text{reduced}}}{T_{\text{total}}}$, where $T_{\text{total}}$ is the total time taken for the task, $T_{\text{reduced}}$ is the time that was reduced.

### 4.3 Success Rate and Efficiency Evaluation

**Success Rate & Task Efficiency.** In Tbl. 1, we present the quantitative results of KARMA and the baselines on the sequence tasks dataset ALFRED-L. For complex tasks, KARMA achieves a 21% task success rate improvement and a 69% reduction in time, which represent a relative improvement of 2.3× and 62.7×, respectively, compared to HELPER (the best-performing baseline in this setting). For composite tasks, KARMA achieves a 43% task success rate improvement and a 68.7% reduction in time, which represent a relative improvement of 1.3x and 3.4x, respectively, compared to CAPEAM (the best-performing baseline in this setting). For simple tasks, KARMA achieves a 42% task success rate improvement and a 61.2% reduction in time, which represent a relative improvement of 1.1× and 2.3×, respectively, compared to HELPER (the best-performing baseline in this setting). It is worth noting that since simple tasks do not require the use of short-term memory, KARMA does not show a significant improvement in task success rate over other baselines.

**Memory Retrieval Accuracy.** We show the accuracy of memory recall in the MRA column of Tbl. 1. Our memory system achieves a recall accuracy that is 2.2× higher for composite tasks compared to complex tasks, as the recall method has certain limitations when instructions contain ambiguous information. We believe this is due to the inherent performance limitations of the commonly used models for semantic matching. For complex tasks, instructions may contain particularly ambiguous semantics, such as "get me a high-calorie food," where even the most advanced semantic matching models perform poorly.

### 4.4 Replacement Policy Evaluation

Fig. 4 illustrates the efficiency of the FIFO policy compared to the W-TinyLFU policy under various configurations of window segment size and main segment size, with a total of 10 memory units. We show the number of consecutive tasks performed by the agent on the x-axis. The y-axis shows the memory hit rate for each memory replacement policy, representing the effectiveness of each policy. Vertical lines of different colors indicate whether the corresponding policy has undergone a warm-up phase. We consider memory to be warmed up when the occupancy rate of the memory units exceeds

Table 1: Evaluation of KARMA and baseline for different categories of tasks in ALFRED-L.

| Methods | Simple Tasks | | | | Composite Tasks | | | | Complex Tasks | | | |
|---|---|---|---|---|---|---|---|---|---|---|---|---|
| | SR | MRA | RE | RT | SR | MRA | RE | RT | SR | MRA | RE | RT |
| LoTa-Bench(Modified) | 0.41 | - | - | - | 0.23 | - | - | - | 0.04 | - | - | - |
| HELPER(Sarch et al., 2023b) | 0.40 | - | 0.251 | 0.263 | 0.21 | - | 0.243 | 0.178 | 0.09 | - | 0.018 | 0.011 |
| CAPEAM(Kim et al., 2023) | 0.35 | - | -0.054 | -0.002 | 0.33 | - | 0.293 | 0.201 | 0.07 | - | 0.012 | 0.008 |
| KARMA | 0.42 | - | 0.582 | 0.612 | 0.43 | 0.93 | 0.902 | 0.687 | 0.21 | 0.42 | 0.867 | 0.690 |

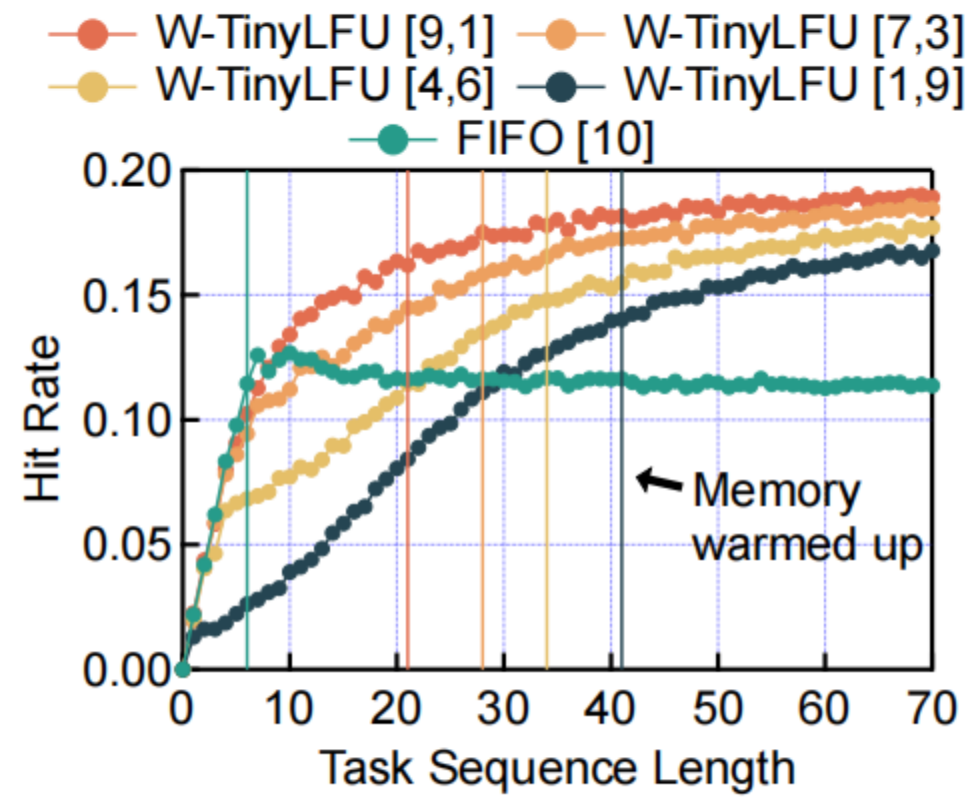

Fig. 4: The memory hit rate of FIFO and W-TinyLFU. [10] means the memory size of FIFO is 10, [9,1] means the memory size of W-TinyLFU is also 10, the main segment is 1, window segment is 9.

95%. After all replacement policies have undergone their warm-up phases, the W-TinyLFU policy with a window segment size of 9 achieves the highest memory hit rate. This indicates that, on the ALFRED-R dataset, a larger window segment size in the W-TinyLFU policy allows for more effective utilization of memory units. For W-TinyLFU, a larger window size typically covers a broader time range, capturing more memory units that are likely to be frequently recalled. These memory units have a high probability of being reused in the task sequence, thereby increasing the memory hit rate.

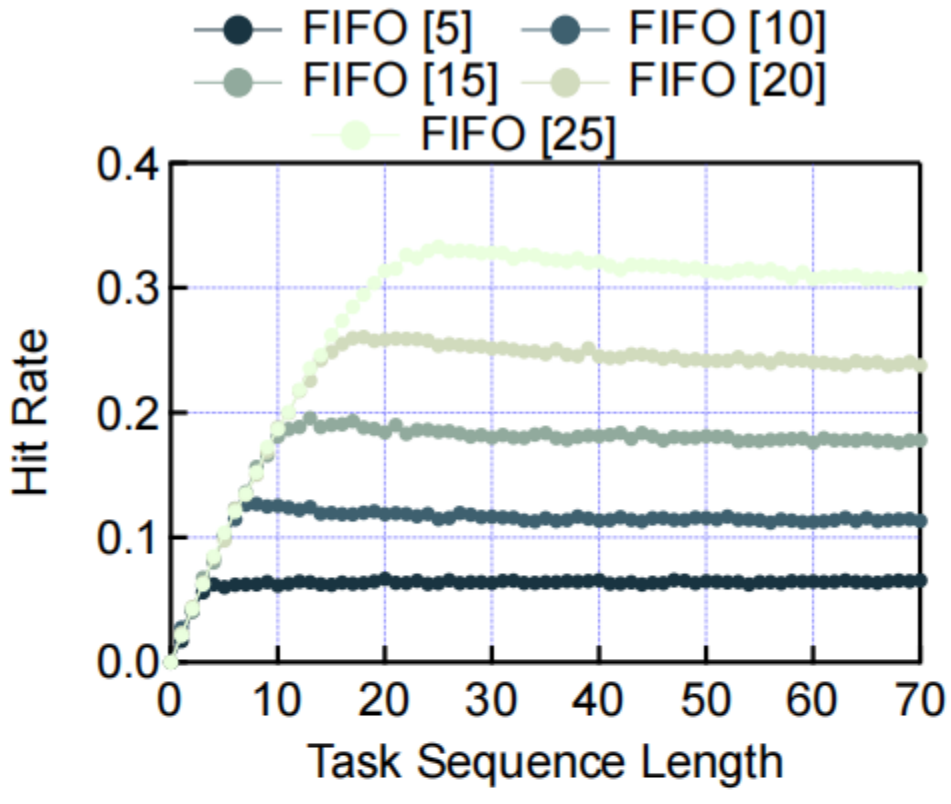

Fig. 5: Evaluation on different FIFO sizes. [10] means the memory is with size equals to 10.

Fig. 5 illustrates the memory hit rate of FIFO policy with different numbers of memory units, with x-axis represents the number of tasks. As expected, larger memory size brings higher hit rate, the memory hit rate with 25 memory units is $4.6\times$ higher than with only 5 memory units. Similar results can be extracted through Fig. 6, where memory hit rate with 25 memory units is $3.9\times$ higher than with only 5 memory units.

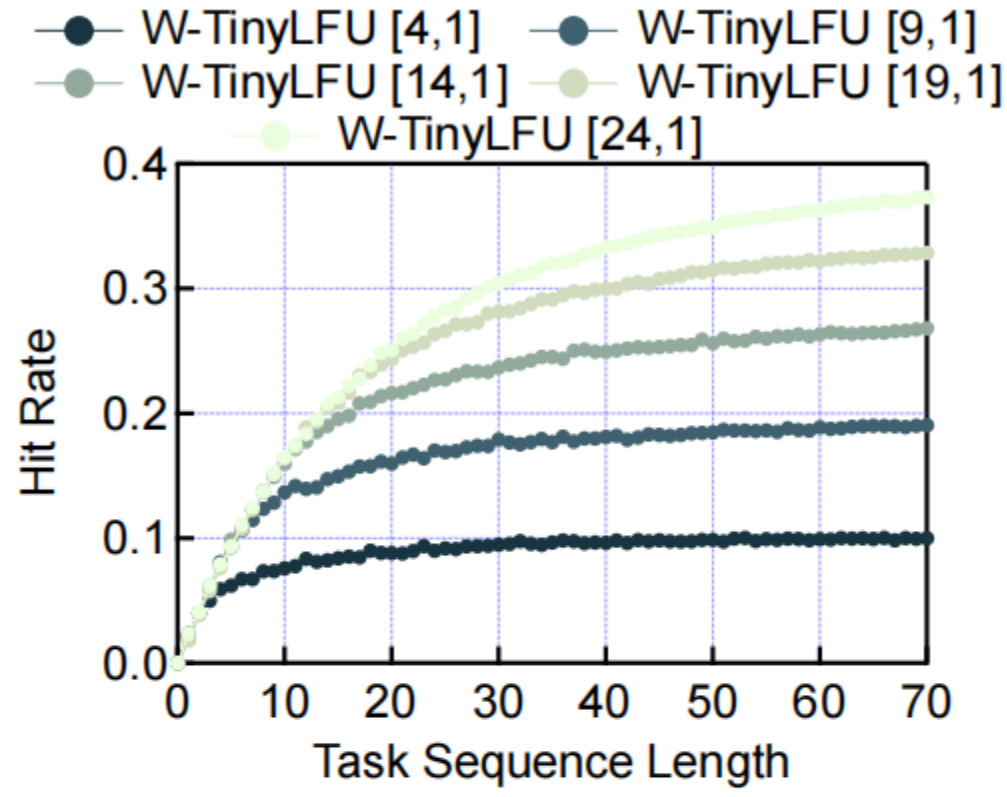

Fig. 6: Evaluation on W-TinyLFU configurations. [9,1] means the memory size of wTinyLFU is 10, the main segment is 1, window segment is 9.

In Fig. 7, we illustrate the impact of memory hit rate on the efficiency of task execution. The x-axis shows the memory hit rate of the W-TinyLFU policy with a window segment size of 9 and a main segment size of 1. The y-axis displays the proportion of reduced exploration. We demonstrate that the memory hit rate and the proportion of reduced exploration are linearly correlated. This means that increasing the memory hit rate enhances the agent's task execution efficiency. A higher memory hit rate signifies more efficient use of memory units. This enhances the agent's ability to recall relevant information, reducing the amount of action code needed for task execution, and ultimately improving overall task performance.

### 4.5 Ablation Study

In Tbl. 2, we evaluate the performance of KARMA after removing short-term memory or long-term memory. The removal of short-term memory significantly affected the agent's ability to handle com-plex and composite tasks, with the success rate dropping by $1.9\times$ and $4.2\times$, respectively. However, this did not greatly impact the agent's task execution efficiency, which decreased only by $1.2\times$ and $1.1\times$. On the other hand, removing long-term memory had a notable impact on task execution efficiency, with the RT decreasing by $2.7\times$, but its effect on success rate was less pronounced, with SR only dropping by $1.2\times$.

Table 2: Ablation Study.

| Methods | Simple Tasks | | | | Composite Tasks | | | | Complex Tasks | | | |
|---|---|---|---|---|---|---|---|---|---|---|---|---|
| | SR | MRA | RE | RT | SR | MRA | RE | RT | SR | MRA | RE | RT |
| LoTa-Bench(Modified) | 0.41 | - | - | - | 0.23 | - | - | - | 0.04 | - | - | - |
| KARMA(w/o long term memory) | 0.40 | - | 0.011 | 0.002 | 0.35 | 1 | 0.329 | 0.210 | 0.12 | 0.43 | 0.021 | 0.013 |
| KARMA(w/o short term memory) | 0.44 | - | 0.573 | 0.605 | 0.22 | - | 0.774 | 0.624 | 0.05 | - | 0.784 | 0.654 |
| KARMA | 0.42 | - | 0.582 | 0.612 | 0.43 | 0.93 | 0.902 | 0.687 | 0.21 | 0.42 | 0.867 | 0.690 |

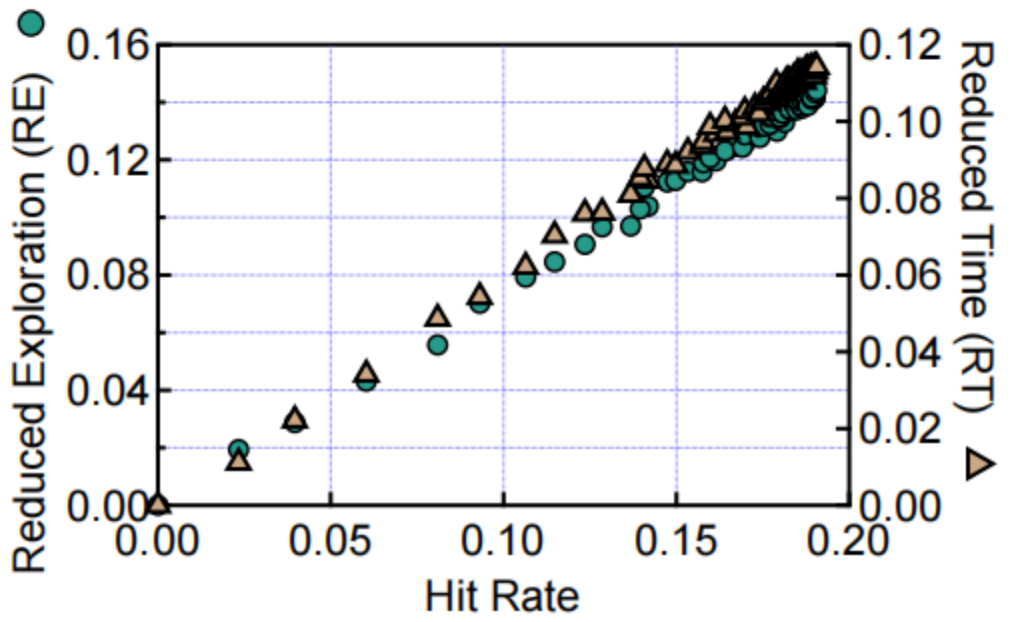

Fig. 7: The impact of memory hit rate on the agent's task execution efficiency.

In summary, short-term memory plays a key role in improving task success rates, while long-term memory has a greater impact on task efficiency. Long-term memory retains 3D scene maps representing the environment, helping to reduce the action code generated by the LLM during task planning, thereby enhancing task execution efficiency. Meanwhile, short-term memory stores information about recently used objects, ensuring that these objects or relevant details are readily accessible for future tasks.

### 4.6 Real-world deployment

We deploy KARMA on a mobile manipulation robot consisting of a UR3 robotic arm and a six-wheeled chassis to demonstrate KARMA's ability to store and retrieve memory, enhancing the LLM's capability for planning long-sequence tasks in real-world environments. For the robot's navigation and obstacle avoidance, we utilize Google Cartographer for simultaneous localization and mapping (SLAM). The camera mounted on the robotic arm's gripper was used to detect objects, feeding the input into LangSAM(Kirillov et al., 2023) for segmentation and semantic matching to locate the object to be grasped. And then AnyGrasp(Fang et al., 2023) generates the most suitable grasping position and plans the arm's motion path.

## 5 Conclusion

In this paper, we explore the potential of enhancing embodied AI agents by integrating external long-and-short term memory systems. Through the implementation of a customized memory system, recall mechanism, and replacement policy, we demonstrate significant improvements over state-of-the-art embodied agents that also utilize memory. Specifically, our memory-augmented AI agent achieves success rates that are $1.3\times$ higher in composite tasks and $2.3\times$ higher in complex tasks. Additionally, task execution efficiency is improved by $3.4\times$ in composite tasks and an impressive $62.7\times$ in complex tasks.

This memory system streamlines the transition from simulation to real-world robotic applications, allowing long- and short-term memory storage and recall methods to be seamlessly integrated into task planning for real robotic systems(Sun et al., 2024).

## 6 Limitations

**Ideal Simulation Environments.** In this work, all evaluations are performed under ideal simulation environments, free from interruptions by other agents or humans. However, this ideal situation is not reflective of real life. Although this paper includes extensive experiments, it lacks evaluation of how the memory system will behave in real-world scenarios. Specifically, the number of objects in the real world will significantly increase compared to a simulation environment, making the effectiveness of recall and replacement mechanisms crucial to final performance. Additionally, we have not tested the system's response to intentional disturbances by humans. These factors constitute the primary limitation of this paper.

**Lack of Biological Theory.** Although effective, the current design of the memory system is analogous to the memory systems of existing computing platforms. For instance, the concept of short-term memory and its replacement can be found in cache design. However, human memory may not function in this manner. This work borrows terminology from human memory yet lacks theoretical support from a biological perspective, which constitutes its second limitation.

**Open-loop Planning.** In this work, all memory operations and planning are open-loop, meaning there is no feedback. However, in most robot system designs, feedback is necessary. For example, if the memory is incorrect, there is no mechanism designed for eviction or updating. The lack of feedback constitutes the third limitation of this paper.

# Supplementary Material

## A Prompts

In Fig. 8, we provide a prompt template that integrates both long-term and short-term memory, specifically designed to enhance the capabilities of LLMs in planning long-sequence tasks.

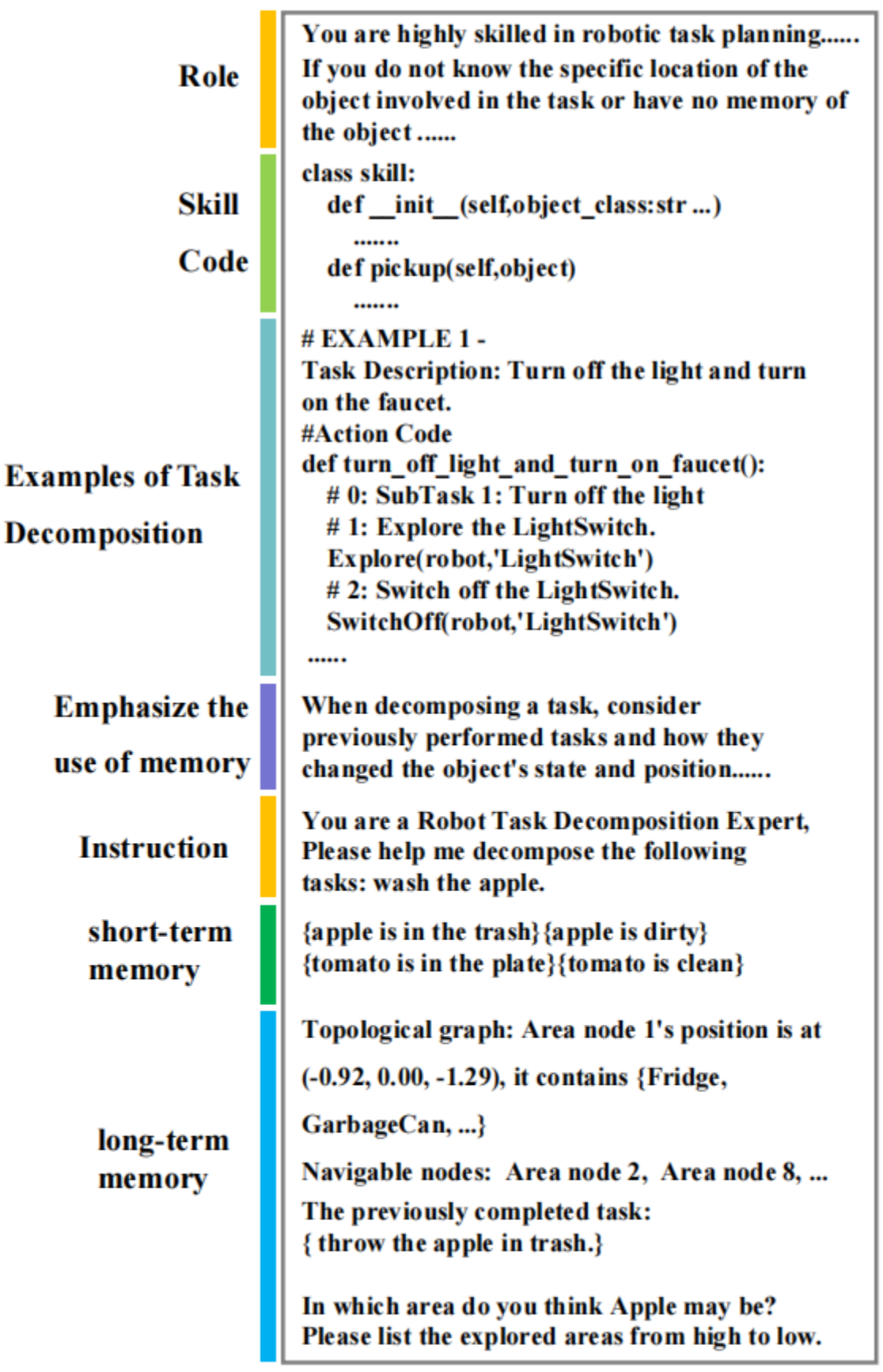

Fig. 8: Our prompt template for LLM encompasses several key elements: the role of LLM, the skill API, examples of task decomposition, an emphasis on the importance of memory, natural language instruction, and the structured recall of both short-term and long-term memory.

## B More Details on Short-Term

We present the contents stored in short-term (Listing1) during task execution . In Listing1, we present the text and image stored in short-term memory after executing the sequential tasks of washing a potato and placing it on the countertop, washing a tomato and placing it on the countertop, putting bread on the countertop, and throwing the knife in the trash. In short-term memory, the "objectId" is a unique identifier for each object that remains constant over time. This identifier is used to determine if the object is the same before and after memory updates. The "position" records the current location of the object after the agent's interaction or the location of objects the agent has encountered during task execution. The "imagePath" stores images of objects captured by the agent, which are used for subsequent analysis by the Vision-Language Model (VLM).

In Fig. 9, we present the image of bread captured by the agent after executing the task of putting bread on the countertop. This image is stored at ”$/short_term/images/Bread.jpg$”.

Listing 1: The detailed content of short-term memory during task execution.

```
short_term_memory=[
    {
        "objectType": "Tomato",
        "position": {
            "x": 0.9792354106903076,
            "y": 1.7150063514709473,
            "z": -2.606173276901245
        },
        "objectId": "Tomato
    |-00.39|+01.14|-00.81"
        "imagePath": "/short_term/images
    /Tomato.jpg"
    },
    {
        "objectType": "Apple",
        "position": {
            "x": 1.0981664657592773,
            "y": 0.9569252133369446,
            "z": -2.4071836471557617
        },
        "objectId": "Apple
    |-00.47|+01.15|+00.48"
        "imagePath": "/short_term/images
    /Apple.jpg"
    },
    {
        "objectType": "DishSponge",
        "position": {
            "x": -1.8567615747451782,
            "y": 0.1449012756347656 2,
            "z": -1.619217514991760 3
        },
        "objectId": "DishSponge
    |-01.94|+00.75|-01.71"
        "imagePath": "/short_term/images
    /DishSponge.jpg"
    },
    {
        "objectType": "Potato",
        "position": {
            "x": 1.09816658496856 7,
            "y": 0.9390283823013306,
            "z": -2.253550529479980 5
        },
        "objectId": "Potato
```

```
    |-01.66|+00.93|-02.15"
        "imagePath": "/short_term/images
    /Potato.jpg"
    },
    {
        "objectType": "Book",
        "position": {
            "x": -1.35060715675354,
            "y": 1.1669094562530518,
            "z": 1.970085859298706
        },
        "objectId": "Book
    |+00.15|+01.10|+00.62"
        "imagePath": "/short_term/images
    /Book.jpg"
    },
    {
        "objectType": "Bread",
        "position": {
            "x": 0.9692967534065247,
            "y": 0.9761490225791931,
            "z": -2.330367088317871
        },
        "objectId": "Bread
    |-00.52|+01.17|-00.03"
        "imagePath": "/short_term/images
    /Bread.jpg"
    },
    {
        "objectType": "Knife",
        "position": {
            "x": -2.0168256759643555,
            "y": 0.24547088146209717,
            "z": 2.1725265979766846
        },
        "objectId": "Knife
    |-01.70|+00.79|-00.22"
        "imagePath": "/short_term/images
    /Knife.jpg"
    },
    {
        "objectType": "Lettuce",
        "position": {
            "x": -1.6119909286499023,
            "y": 0.9801480174064636,
            "z": -0.6989647150039673
        },
        "objectId": "Lettuce
    |-01.81|+00.97|-00.94"
        "imagePath": "/short_term/images
    /Lettuce.jpg"
    }
]
```

## C More Details on ALFRED-L

ALFRED-L includes three types of tasks: simple tasks, composite tasks, and complex tasks. These tasks are adapted from the original ALFRED dataset. In ALFRED-L, placing an object inside the fridge was deemed successful when the object is in the fridge. We enhanced this by adding a subgoal "INSIDE(Fridge): 1" to ensure the object is correctly placed inside fridge. For tasks like "wash an apple" in ALFRED-L, the goal conditions involve the apple being rinsed in the sink.

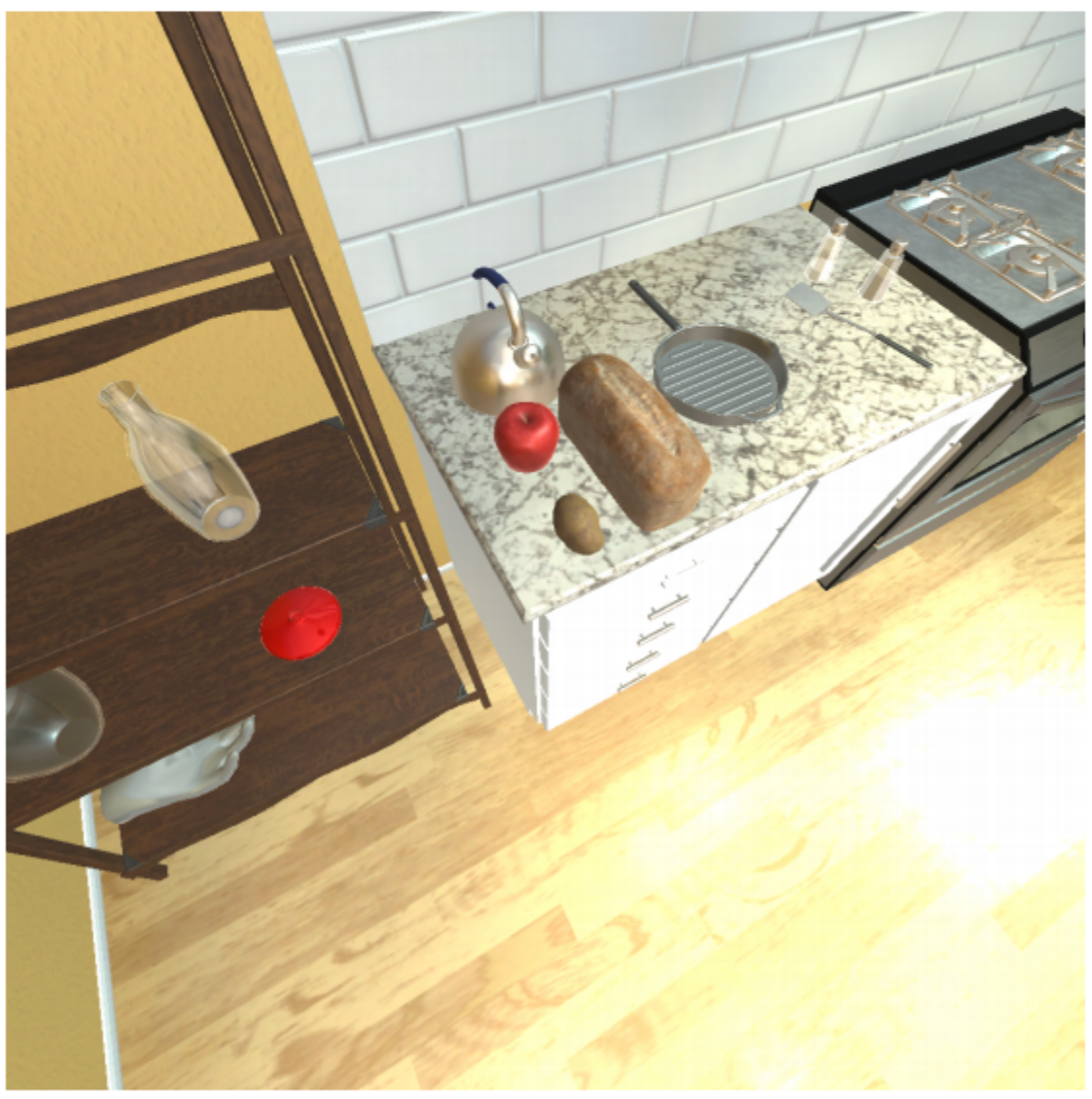

Fig. 9: The image stores at "/short_term/images/Bread.jpg" was captured after the task of putting bread on the countertop was executed.

This requires multiple conditions to be met, such as "INSIDE(apple, sink): 1", "TOGGLEON(Faucet): 1", and "State(apple, clean): 1". Examples of instructions and goal conditions from the dataset are shown in Tbl. 3.

## D Skill API and Action Code

We provide detailed skill APIs and their corresponding action codes in the Listing2.

## E LANGUAGE MODELS

Tbl. 4 lists the language models used in experiments and outlines their core functions.

## F Details of image analysis in short-term memory

In Fig. 10, we present the prompt used to analyze images stored in short-term memory by the Vision-Language Model (VLM). The text highlighted in blue, [Image], represents the placeholder that will be filled with an image, while [task] will be replaced with the actual instruction. We employed a step-by-step Chain of Thought approach to guide the VLM in identifying the relevant objects and their corresponding states.

## G An example result of KARMA on the ALFRED-L dataset

In Fig. 11, we present images of the agent performing tasks in the AI2-THOR simulator.

Listing 2: Full Skill API and Action CODE used in the prompts.

```
def GoToObject(robots, dest_obj):
    # Navigate to the object.

    # If agent knows the location of object, the agent can use this function to
    navigates to the object.
    # If agent does not know the location of object, the robot should use the
    Explore(robots, dest_obj) to find the object.

    # The function captures only those objects that are within the agent's line of
    sight.

    # Example:
    # <Instruction> Go to the apple(The memory contains the location of apple).
    # Python script:
    # GoToObject(robot,'Apple')
    pass

def PickupObject(robot, pick_obj):
    # pickup the object.
    # The function captures only those objects that are within the agent's line of
    sight.

    # Example:
    # <Instruction> Go get the apple on the kitchen counter.
    # Python script:
    # Explore(robot,'CounterTop')
    # GoToObject(robot,'CounterTop')
    # PickupObject(robot,'CounterTop')
    pass

def PutObject(robot, put_obj, recp):
    # puts the current interactive object held by the agent in the designated
    location.
    # This function assumes the object is already picked up.

    # Example:
    # <Instruction> put the apple on the Sink.
    # Python script:
    # Explore(robot,'Sink')
    # GoToObject(robot,'Sink')
    # PutObject(robot,'Sink')
    pass

def SwitchOn(robot, sw_obj):
    # Turn on a switch.

    # Example:
    # <Instruction> Turn on the light.
    # Python script:
    # SwitchOn(robot,'LightSwitch')
    pass

def SwitchOff(robot, sw_obj):
    # Turn off a switch.

    # Example:
    # <Instruction> Turn off the light.
    # Python script:
    # SwitchOn(robot,'LightSwitch')
    pass

def OpenObject(robot, sw_obj):
    # Open the interaction object.
    # This function assumes the object is already closed and the agent has already
    navigated to the object.
    # Only some objects can be opened. Fridges, cabinets, and drawers are some
    example of objects that can be closed.
```

```
    #Example:
    # <Instruction> Get the apple in the fridge.
    # Python script:
    # Explore(robot,'Fridge')
    # GoToObject(robot,'Fridge')
    # OpenObject(robot,'Fridge')
    # PickupObject(robot,'apple')
    pass

def CloseObject(robot, sw_obj):
    # Close the interaction object.
    # This function assumes the object is already open and the agent has already
    navigated to the object.
    # Only some objects can be closed. Fridges, cabinets, and drawers are some
    example of objects that can be closed.
    pass

def BreakObject(robot, sw_obj):
    # Break the interactable object.
    pass

def SliceObject(robot, sw_obj):
    # Slice the interactable object.
    # Only some objects can be sliced. Apple, tomato, and bread are some example of
    objects that can be sliced.

    #Example:
    # <Instruction> Slice an apple.
    # Python script:
    # Explore(robot,'Knife')
    # GoToObject(robot,'Knife')
    # PickupObject(robot,'Knife')
    # Explore(robot,'Apple')
    # GoToObject(robot,'Apple')
    # SliceObject(robot,'Apple')
    pass

def ThrowObject(robot, sw_obj):
    # Throw away the object.
    # This function assumes the object is already picked up.
    pass

def Explore(robot, sw_obj, position):
    # Explore the environment in the given sequence of locations until the target
    object becomes visible in the robot's field of view.
    pass

def ToggleOn(robot, sw_obj):
    # Toggles on the interaction object.
    # This function assumes the interaction object is already off and the agent has
    navigated to the object.
    # Only some landmark objects can be toggled on. Lamps, stoves, and microwaves
    are some examples of objects that can be toggled on.

    # Example:
    # <Instruction> Turn on the lamp.
    # Python script:
    # Explore(robot,'Lamp')
    # GoToObject(robot,'Lamp')
    # ToggleOn(robot,'Lamp')
    pass

def ToggleOff(robot, sw_obj):
    # Toggles off the interaction object.
    pass
```

Table 3: Task types and samples for each type in the ALFRED-L dataset.

| **Task Type** | **Goal Condition** | **Instruction** |
|---|---|---|
| Simple Tasks | ON(apple, plate): 1,<br>INSIDE(apple, sink): 1,<br>TOGGLEON(faucet): 1,<br>STATE(apple, clean): 1,<br>HOLDON(robot, frying pan): 1 | place potato on the plate → wash an apple → get a frying pan. |
| Composite Tasks | STATE(tomato, sliced): 1,<br>INSIDE(knife, garbageCan): 1,<br>TOGGLEON(Faucet): 1,<br>ON(tomato, plate): 1 | slice a tomato → throw the knife in the trash → place the tomato on the plate. |
| Complex Tasks | INSIDE(tomato, sink): 1,<br>TOGGLEON(faucet): 1,<br>STATE(tomato, clean): 1,<br>INSIDE(potato, sink): 1,<br>TOGGLEON(faucet): 1,<br>STATE(potato, clean): 1,<br>STATE(bread, sliced): 1,<br>INSIDE(bread, fridge): 1,<br>ON(tomato, plate): 1 | wash a tomato → wash a potato → slice a bread → put the bread in the fridge → place the clean, red food on the plate. |

Table 4: List of language models used in the experiments and their respective roles.

| **Language Model** | **Role** | **Function** |
|---|---|---|
| OpenAI GPT-4o | VLM | Analyzes the state of objects within the image of short-term memory. |
| OpenAI GPT-4o | LLM as Planner | Task decomposition. |
| OpenAI text-embedding-3-large | Embedding Model | Recalls memory units. |

**<System Rolo> As an image analysis expert, your task is to infer the state of objects in the image through step-by-step reasoning.**

**<User Role>**

**1.Provide a detailed description of this image[Image].**

**2.From the given task[Task], extract the relevant content from the first step's image description that pertains to the mentioned objects.**

**3.Based on the object descriptions extracted in the second step, match each object to one of the following states: heated, cooked, sliced, cleaned, dirty, filled, used up, off, on, opened, closed, none.**

**4.Summarize the results from step three in the following format: object: state.**

Fig. 10: The prompt template for GPT-4, utilizing a step-by-step approach to guide VLM in identifying the relevant objects and their corresponding states.

**Instruction: wash an tomato and place it on the countertop ➡ find an apple and place it on the countertop ➡ slice the clean tomato**

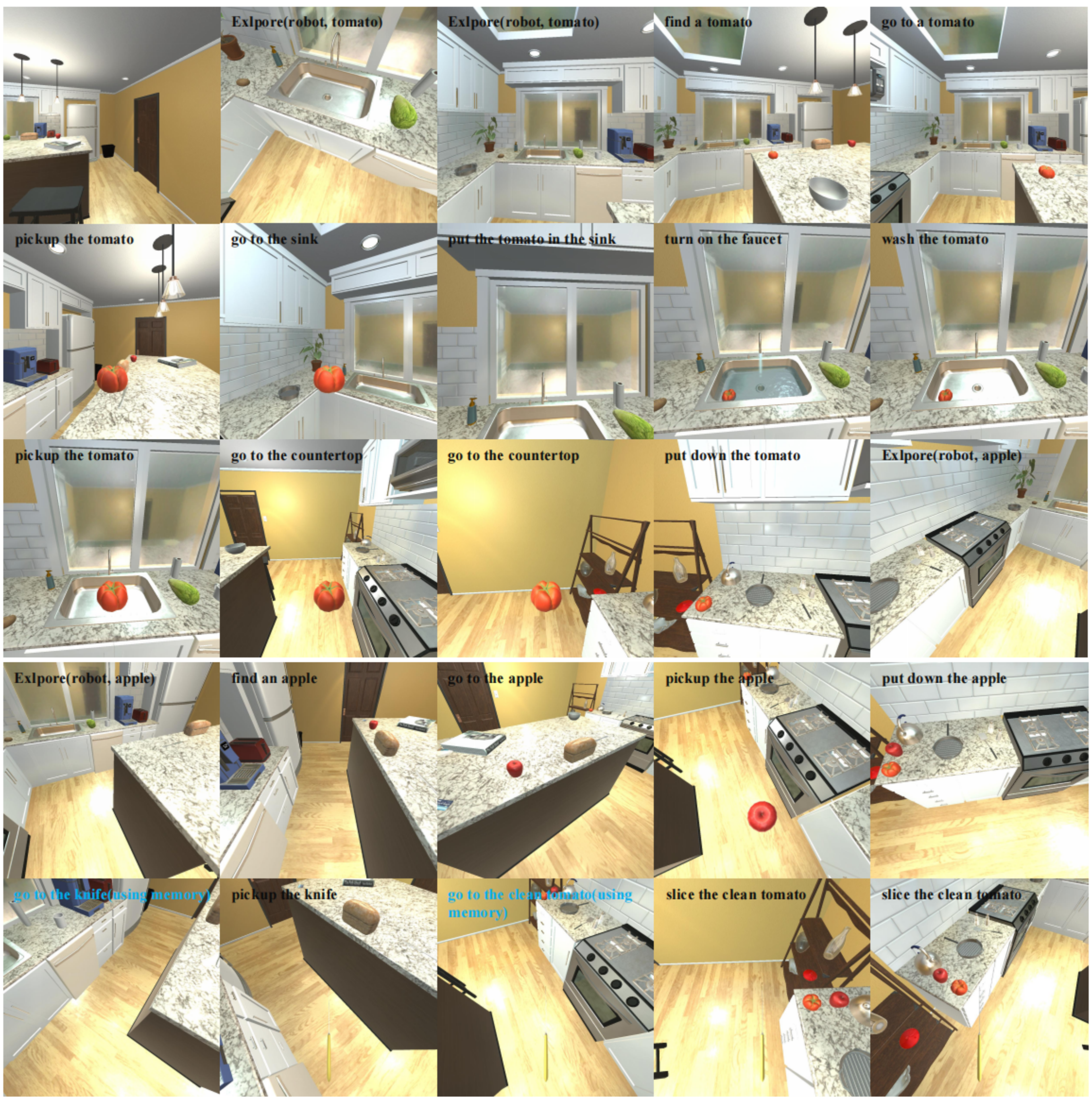

Fig. 11: An example result of KARMA on the ALFRED-L dataset.